%% file: neurips_2025.tex
\documentclass{article}

\PassOptionsToPackage{numbers,sort&compress}{natbib}

\usepackage[dblblindworkshop,final]{neurips_2025}

\workshoptitle{MATH-AI}
\usepackage{siunitx}

\usepackage[utf8]{inputenc}
\usepackage[T1]{fontenc}
\usepackage{textcomp} % safer monospaced punctuation, etc.
\usepackage{newunicodechar}
\newunicodechar{≠}{\ensuremath{\neq}}
\newunicodechar{≤}{\ensuremath{\le}}
\newunicodechar{≥}{\ensuremath{\ge}}
\newunicodechar{→}{\textrightarrow}
\newunicodechar{←}{\textleftarrow}
\newunicodechar{α}{\ensuremath{\alpha}}

\newunicodechar{τ}{\ensuremath{\tau}}
\newunicodechar{σ}{\ensuremath{\sigma}}

\newunicodechar{φ}{\ensuremath{\varphi}}
\newunicodechar{ϕ}{\ensuremath{\phi}}

\usepackage[skins,breakable]{tcolorbox}
\usepackage{fvextra} % enhanced verbatim: \VerbatimInput with line breaks

\newcommand{\ShowMD}[2]{%
  \begin{tcolorbox}[breakable,title={#1}]
    \VerbatimInput[
      breaklines=true,    % wrap long lines
      breakanywhere=true, % allow breaks inside long tokens
      obeytabs=true, tabsize=2,
      fontsize=\footnotesize
    ]{#2}%
  \end{tcolorbox}%
}

\usepackage{hyperref}       % hyperlinks
\usepackage{url}            % simple URL typesetting
\usepackage{booktabs}       % professional-quality tables
\usepackage{amsfonts}       % blackboard math symbols
\usepackage{nicefrac}       % compact symbols for 1/2, etc.
\usepackage{microtype}      % microtypography
\PassOptionsToPackage{table}{xcolor}
\usepackage{xcolor}         % colors
\usepackage{colortbl}    
\usepackage{makecell} 
\usepackage{svg}
\usepackage{tikz}
\usepackage{array}
\usetikzlibrary{arrows.meta,positioning}
\usepackage[edges]{forest}
\usepackage{tabularx}

\input{math_commands.tex}

\tcbuselibrary{listings}
\usepackage{graphicx}
\usepackage{algorithm}
\usepackage{algorithmicx}
\usepackage{algpseudocode}
\usepackage{float}
\usepackage{placeins}
\usepackage{caption}

\usepackage{tasks}
\settasks{label=(\arabic*),label-format={\bfseries},after-item-skip=0pt,column-sep=1em}

\usepackage{wrapfig}
\usepackage{subcaption}
\definecolor{lightcream}{HTML}{F5EFE6}
\definecolor{warmbeige}{HTML}{E8DFCA}
\definecolor{mediumblue}{HTML}{6D94C5}
\definecolor{lightblue}{HTML}{CBDCEB}
\definecolor{darkercream}{HTML}{E6D5C0}
\definecolor{darkerbeige}{HTML}{D4C4A0}
\definecolor{softblue}{HTML}{A8C0E0}

\definecolor{headergray}{HTML}{F2F2F2}
\definecolor{bestcell}{HTML}{D9F2D9}

\newcommand{\best}[1]{\cellcolor{bestcell}{\bfseries #1}}

\title{CombiGraph-Vis: A Curated Multimodal Olympiad Benchmark for Discrete Mathematical Reasoning}

\author{%
\textbf{Hamed Mahdavi}$^{1}$ \quad \textbf{Pouria Mahdavinia}$^{1}$ \quad \textbf{Alireza Farhadi}$^{4}$ \quad \textbf{Pegah Mohammadipour}$^{1}$ \\
\textbf{Samira Malek}$^{1}$ \quad \textbf{Majid Daliri}$^{3}$ \quad \textbf{Pedram Mohammadipour}$^{4}$ \quad \textbf{Alireza Hashemi}$^{2}$ \\
\textbf{Amir Khasahmadi}$^{5}$ \quad \textbf{Vasant Honavar}$^{1}$ \\[4mm]
$^{1}$Pennsylvania State University \quad
$^{2}$City University of New York \quad
$^{3}$New York University \\
$^{4}$Amirkabir University of Technology \quad
$^{5}$Autodesk
}

\begin{document}

\maketitle

\begin{abstract}
CombiGraph-Vis is a 1,135-problem benchmark for discrete mathematical reasoning spanning 13 domains and three formats (short-answer, multiple-choice, and yes/no). Notably, 35\% of problems include images whose structure is essential for finding solutions. Each problem comes with a verified solution and technique labels, with the entire dataset curated and validated through agentic workflows under human oversight to ensure consistency and fidelity. Evaluations across diverse model families reveal a wide performance range (16\%--78\% accuracy), with particularly sharp drops on image-based problems. For standalone multiple-choice problems, clear gaps emerge between correct-answer accuracy and among-choices accuracy, indicating vulnerability to trap choices. The benchmark emphasizes reasoning over graphs, grids, and other combinatorial objects. We release the dataset, solutions, technique labels, and evaluation code to support research on robust multimodal discrete-math reasoning.~\href{https://github.com/combigraphviz2025/combigraph-viz}{https://github.com/combigraphviz2025/combigraph-viz}
% Place near the top of the Introduction
\begin{figure}[H]
  \centering
  \includegraphics[width=0.815\linewidth]{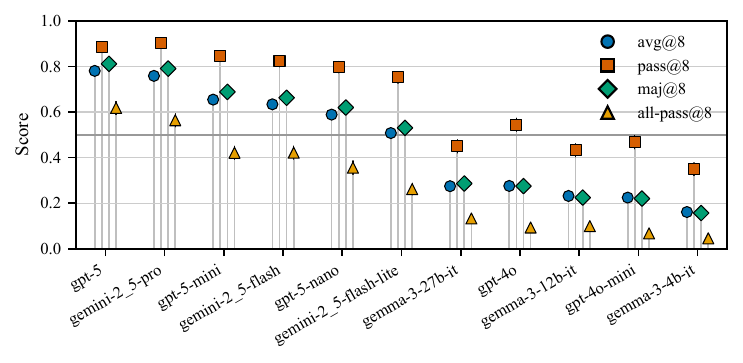}
  \vspace{-0.5em}
  \captionsetup{width=0.81\linewidth, justification=justified, singlelinecheck=false}
  \caption{\textbf{Per-model performance on CombiGraph-Vis.} Across all 1135 problems, the performance of each model is summarized by four evaluation tracks: \texttt{avg@8} (average score), \texttt{pass@8} (any-of-8), \texttt{maj@8} (majority-of-8), and \texttt{all-pass@8} (all-of-8) under our eight-sample chain-of-thought protocol.}
  \label{fig:per-model-all}
  \vspace{-0.75em}
\end{figure}
\end{abstract}

\section{Introduction}
General-purpose math benchmarks increasingly show ceiling effects, limiting their ability to differentiate model capabilities, while multimodal datasets often under-represent discrete mathematics \citep{cobbe2021training,hendrycks2021math}. Recent multimodal efforts broaden coverage but lack the depth needed to assess discrete mathematical reasoning skills \citep{lu2024mathvista,wang2024mathv}. Competition-oriented benchmarks are valuable, but they mostly feature problems crafted to demand elaborate proofs rather than to elicit short answers, even when the problems admit a concrete final answer~\citep{mao-etal-2024-champ,gao2024omnimath}. Multimodal competition collections exist, but they are not focused on discrete mathematics with fine-grained technique analyzes \citep{he-etal-2024-olympiadbench}.

We introduce CombiGraph-Vis, a benchmark of 1135 problems for discrete mathematical reasoning across 13 domains and three formats (short-answer, multiple-choice, yes/no), with substantial visual content (35\% image-tagged problems). Each problem includes a verified solution and technique labels, and the dataset is curated and validated via agentic workflows with human oversight to ensure consistency and fidelity. Problems stress reasoning over graphs, grids, combinatorial constructions, and logic-driven puzzles, yielding short, checkable answers.

Evaluations across diverse model families show strong separation on single-sample accuracy (16\%--78\%), larger drops on image-tagged items, and clear differences between correct-answer accuracy and among-choices accuracy of standalone multiple-choice probelms, indicating susceptibility to distractors. CombiGraph-Vis complements prior resources by centering discrete mathematics, providing verified solutions and technique labels, and enabling targeted analysis of multimodal reasoning. We release the dataset, corrected solutions, labels, and evaluation code to support research on robust multimodal discrete-math reasoning.

\section{CombiGraph-Vis Dataset}
CombiGraph-Vis is a 1135-problem benchmark for discrete mathematical reasoning across 13 domains and three formats (short-answer, multiple-choice, yes/no). About one-third of problems include images. Each problem includes a verified solution and technique labels.

\subsection{Data Collection}
We gathered all of the problems from the Iranian National Olympiad in Informatics first and second round competitions through the years. Formats in the source competitions changed over time, shifted from mainly multiple-choice to also short-answer and yes/no. We collected first-round problems (5–34) and selected second-round sets (24th, 25th, 26th, 30th, 32nd). PDFs were the primary source. We used \href{Opedia.ir}{the official website} of the competition to validate and fill gaps. When questions shared definitions or a setup, we stored the shared text in \texttt{context} for evaluation of context-dependent problems (Figure~\ref{fig:context_example}). We redrew figures when originals were low resolution or contained Persian text. We used an agentic workflow to label multiple-choice problems as standalone or choice-dependent. We show toy examples in the main text (Figure~\ref{fig:examples}), and full examples are in Appendix~\ref{app:examples}.

% (Table moved to Appendix~\ref{app:dataset_stats})

\begin{figure}[t]
\centering
\subcaptionbox{toy illustrative examples.\label{fig:examples}}[0.66\linewidth]{%
\begin{tcolorbox}[
    colback=lightcream,
    colframe=mediumblue,
    colbacktitle=darkercream,
    coltitle=black,
      title={\textbf{Choice-Dependent Problem}},
    fonttitle=\bfseries,
    rounded corners,
    boxrule=0.8pt,
    left=4pt,right=4pt,top=4pt,bottom=4pt
]
    \textbf{Which statement must hold for every tree with $n\ge2$ vertices?}
\begin{enumerate}
        \item It has exactly one cycle.
        \item It has at least two leaves.
        \item Its average degree is at least $2$.
        \item It contains a triangle.
        \item None of the above.
\end{enumerate}
\end{tcolorbox}
  \vspace{0.5em}
\begin{tcolorbox}[
    colback=warmbeige,
    colframe=mediumblue,
    colbacktitle=darkerbeige,
    coltitle=black,
      title={\textbf{Standalone Problem}},
    fonttitle=\bfseries,
    rounded corners,
    boxrule=0.8pt,
    left=4pt,right=4pt,top=4pt,bottom=4pt
]
    \textbf{How many $5$-bit strings contain exactly two $1$s?}
    \begin{tasks}[label=\arabic*.,label-format={\bfseries},after-item-skip=0pt,column-sep=1em](5)
      \task 5
      \task 8
      \task 10
      \task 12
      \task 16
    \end{tasks}
\end{tcolorbox}
}%
\hfill
\subcaptionbox{Agentic validation pipeline.\label{fig:pipeline}}[0.33\linewidth]{%
  \centering
  \IfFileExists{figures/workshop_problem_validation.png}{%
    \includegraphics[width=\linewidth]{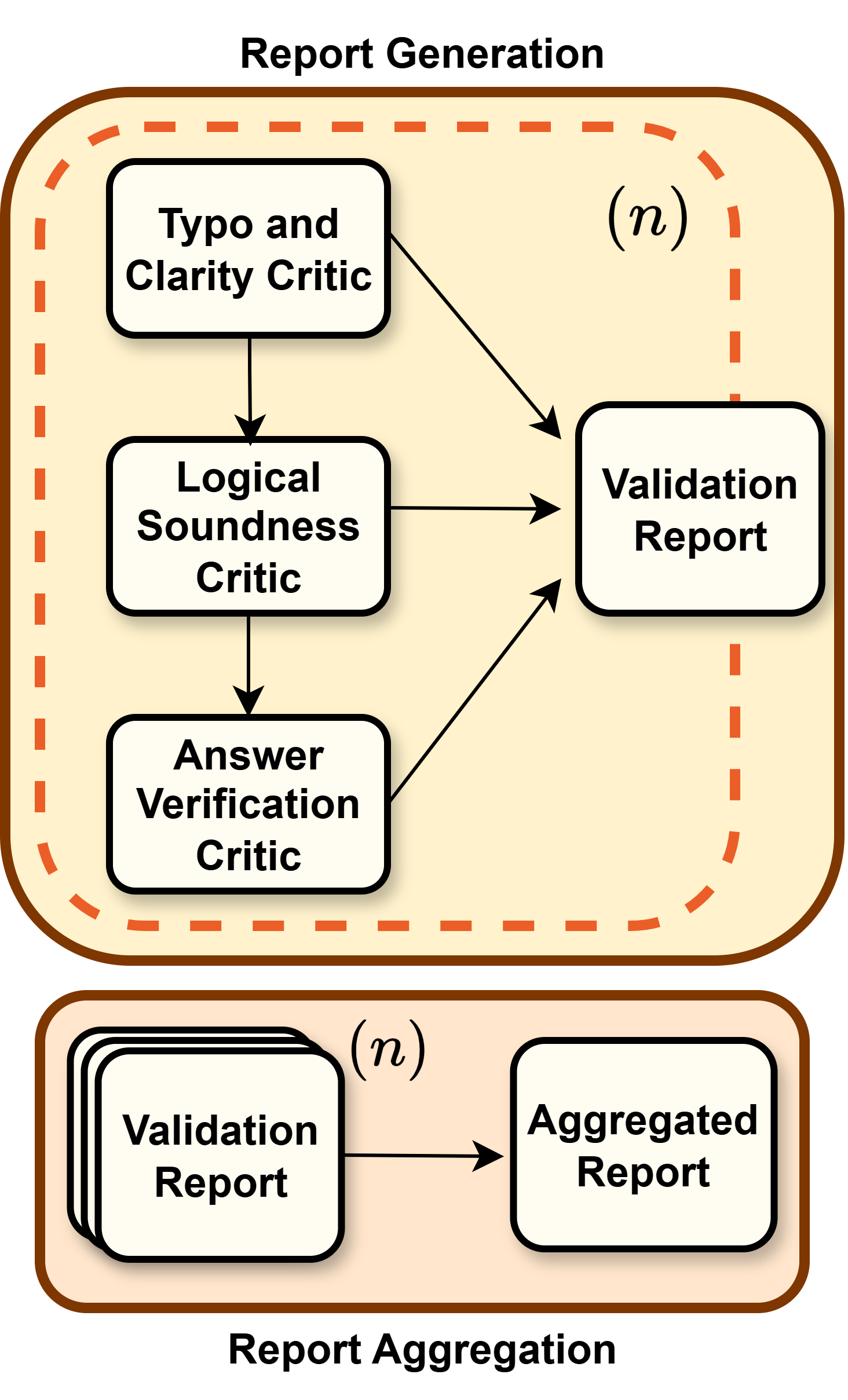}%
  }{%
    \includegraphics[width=\linewidth]{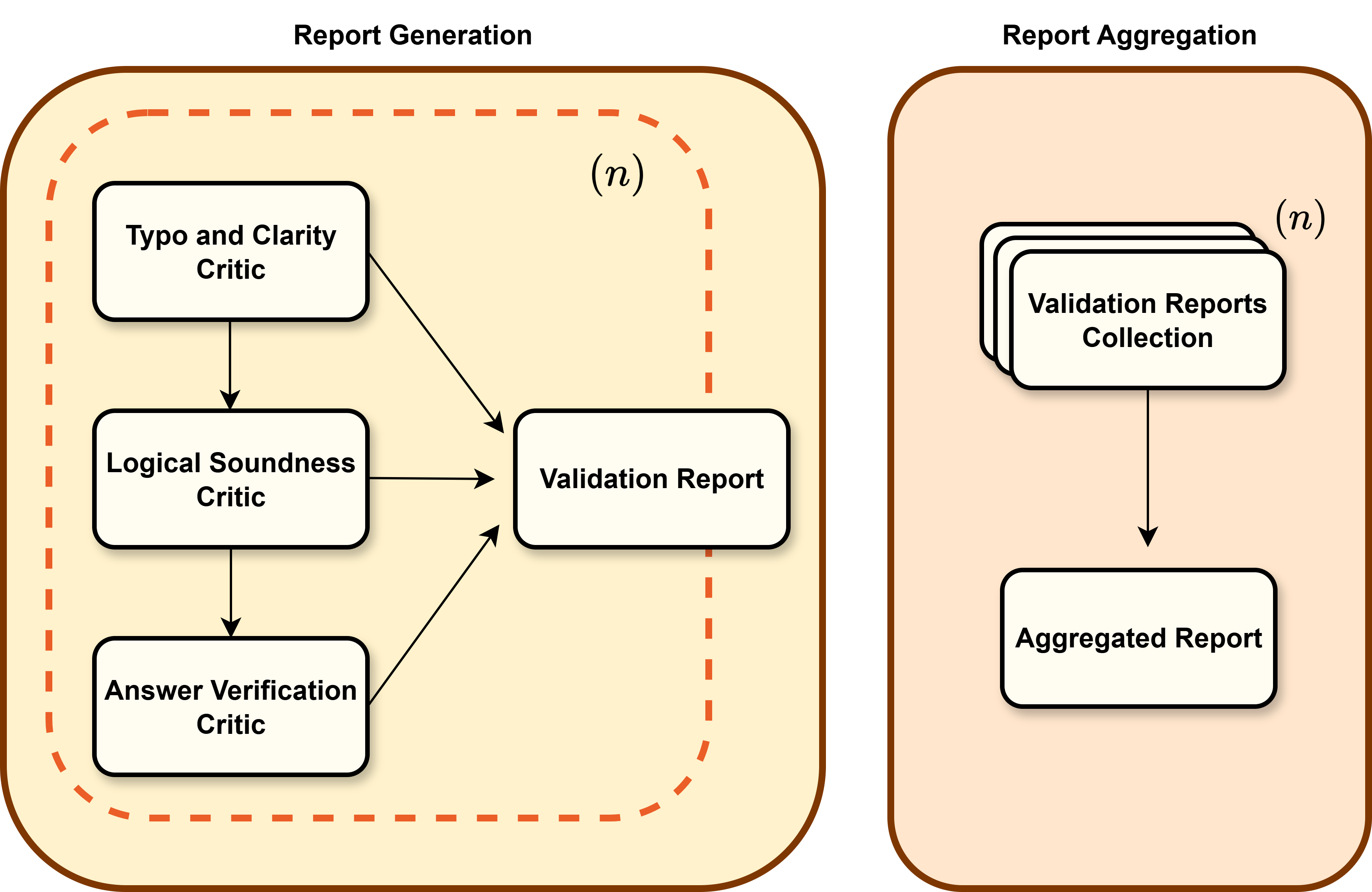}%
  }
}%
\end{figure}

\begin{figure}[t]
\begin{tcolorbox}[
    colback=lightblue,
    colframe=mediumblue,
    colbacktitle=softblue,
    coltitle=black,
    title={\textbf{Context-Dependent Problem (toy example)}},
    fonttitle=\bfseries,
    rounded corners,
    boxrule=0.8pt,
    left=4pt,right=4pt,top=4pt,bottom=4pt
]
\textbf{Context:} A \emph{step sequence} of length $n$ is an integer sequence $(a_1,\dots,a_n)$ with $|a_{i+1}-a_i|=1$ for all $i$. Assume $a_1=0$.

\textbf{Question:} How many step sequences of length $5$ end at $a_5=2$?
\begin{tasks}[label=\arabic*.,label-format={\bfseries},after-item-skip=0pt,column-sep=1em](5)
  \task 2 \task 3 \task 4 \task 5 \task 6
\end{tasks}
\end{tcolorbox}
\caption{Illustrative toy example of a context-dependent problem.}
\label{fig:context_example}
\end{figure}

% (pipeline figure moved into Figure~\ref{fig:examples-and-pipeline} as a subfigure)

\subsection{Data Curation Process Using Agentic Workflows}
We applied agentic workflows with human-in-the-loop to fix existing errors in the dataset during the data curation phase. Our initial analysis identified three distinct error categories with different patterns requiring specialized detection approaches. PDF to markdown conversion errors, translator/annotator errors, and original source errors.

\textbf{First Phase: Problem Validation.} We developed a two-phase validation and correction process using agentic workflows to detect mistakes in problems and solutions. Our first phase uses an agentic workflow that generates validation reports through three specialized critics (Figure~\ref{fig:pipeline}). Each critic examines the complete problem data—including context (if any), text, and answer choices—alongside both Persian and English solutions, the correct option, and final answer. The critics are: \textbf{Typo/Clarity Critic}, \textbf{Logical Soundness Critic}, and \textbf{Final Answer Match}. We run these critic stages three times independently for each problem to generate three validation reports. We then use an aggregator stage that applies majority voting to synthesize the three reports into one structured output report with multiple diagnostic fields. Complete implementation details for the first phase are provided in Algorithm~\ref{alg:problem_validation} (Appendix~\ref{app:implementation}). To filter problematic cases, we use the Overall Error Severity taxonomy with five categories: No issues (1), Minor issues (2), Moderate issues (3), Major issues (4), and Critical failure (5). We checked the generated reports for a handful of cases and detected systematic patterns where problems flagged with "major issues" typically contained only minor typos, while those marked "critical failure" often had single correctable errors. We selected all cases with severity scores above 1 for the second validation and error correction phase, accepting this conservative threshold to minimize false negatives while managing the high false positive rate we observed.

\textbf{Second Phase: Automated Error Resolution.} Many first‑phase flags were not true source errors (e.g., parsing/formatting artifacts, translation slips, or misreads of brief official solutions), so we add a second phase to separate these from genuine issues and apply targeted fixes. The workflow first classifies each case (from aggregated first‑phase reports) as: (i) a pipeline parsing/conversion issue, (ii) a potential original‑source error, or (iii) an image‑understanding issue. Pipeline issues receive surgical edits to every existing field except the original Persian problem and solution, followed by validation. Potential original‑source cases undergo a solution‑engagement pass that expands the brief solution, tries to understand the detected issue and attempts to reclassifies the case; minor, fixable cases proceed with automated edits, while major cases or image-understanding issue are flagged for later human review. The detailed algorithms for this workflow can be found in ~\ref{alg:error_detection} and~\ref{alg:automated_fixing} (Appendix~\ref{app:implementation}).

\section{Results}
Across all evaluation settings, we observe clear separations between model families, with top-tier models achieving strong but far from saturated accuracy, mid-tier models trailing substantially, and lightweight/open-weight models far behind. Accuracy drops on image-tagged items compared to text-only items, revealing persistent gaps in visual mathematical understanding. Analysis of a standalone data subset shows that models are often lured by wrong choices deliberately crafted to make competition settings more challenging.

\noindent\textbf{Overall Performance}
Our results are summarized in Table~\ref{tab:overall_and_mc_combined} (cf. Figure~\ref{fig:per-model-all}). Top-tier models achieve \texttt{avg@8} accuracy of approximately 75--78\%, whereas mid-tier and lightweight/open-weight models lag by 20--40 percentage points across evaluation settings. This broad dispersion persists across formats and modalities, confirming that CombiGraph-Vis is not saturated: even the strongest models leave substantial headroom while weaker models remain far from ceiling. The per-model tracks (avg@8, pass@8, maj@8, all-pass@8) further reinforce clear separations among model families.

\begin{table}[H]
 \centering
 \scriptsize
 \setlength{\tabcolsep}{3pt}
 \caption{Per-model accuracy on CombiGraph-Vis (\texttt{avg@8}). Columns report overall, image slices, multiple-choice (Choice-Dep.), yes/no, and the second-round subset. The three rightmost columns quantify multiple-choice behavior in standalone setting: MC Standalone, Among-Choices, and $\Delta$ = (Among-Choices $-$ Standalone), a proxy for distractor susceptibility. Bold marks the best score in each column.}
 \label{tab:overall_and_mc_combined}
 \begin{tabularx}{\linewidth}{>{\bfseries\raggedright\arraybackslash}X
                             S[table-format=2.1]
                             S[table-format=2.1]
                             S[table-format=2.1]
                             S[table-format=2.1]
                             S[table-format=2.1]
                             S[table-format=2.1]
                             S[table-format=2.2]
                             S[table-format=2.2]
                             S[table-format=2.2]}
 \toprule
 \rowcolor{headergray}
{Model} & {All} & {Img Yes} & {Img None} & {Choice-Dep.} & {Yes/No} & {Second Round} & {MC Standalone} & {Among-Choices} & {$\Delta$} \\
\midrule
\rowcolor{white}
 gemini-2\_5-flash & 63.4 & 50.9 & 70.3 & 56.9 & 74.1 & 50.4 & 63.45 & 83.73 & 20.28 \\
\rowcolor{lightgray}
 gemini-2\_5-flash-lite & 50.8 & 33.8 & 60.2 & 50.6 & 66.4 & 30.2 & 49.19 & 73.08 & 23.89 \\
\rowcolor{white}
 gemini-2\_5-pro & 75.8 & 66.9 & \best{80.8} & 72.9 & \best{81.9} & 71.6 & 75.75 & 90.04 & \best{14.29} \\
\rowcolor{lightgray}
 gemma-3-12b-it & 23.2 & 17.5 & 26.3 & 31.1 & 28.3 & 13.7 & 21.25 & 65.41 & 44.17 \\
\rowcolor{white}
 gemma-3-27b-it & 27.5 & 20.1 & 31.6 & 38.5 & 32.4 & 12.6 & 24.99 & 70.44 & 45.45 \\
\rowcolor{lightgray}
 gemma-3-4b-it & 16.1 & 12.1 & 18.4 & 15.9 & 40.6 & 9.7 & 13.59 & 57.46 & 43.87 \\
\rowcolor{white}
 gpt-4o & 27.6 & 20.4 & 31.6 & 31.4 & 49.9 & 15.9 & 24.55 & 64.09 & 39.55 \\
\rowcolor{lightgray}
 gpt-4o-mini & 22.5 & 16.9 & 25.5 & 25.2 & 50.8 & 14.6 & 18.97 & 60.42 & 41.45 \\
\rowcolor{white}
 gpt-5 & \best{78.0} & \best{68.2} & \best{83.5} & \best{81.2} & 75.7 & \best{75.6} & \best{77.74} & \best{92.03} & \best{14.29} \\
\rowcolor{lightgray}
 gpt-5-mini & 65.4 & 53.9 & 71.8 & 69.0 & 37.4 & 59.9 & 67.82 & 85.40 & 17.58 \\
\rowcolor{white}
 gpt-5-nano & 58.9 & 43.5 & 67.5 & 55.4 & 44.4 & 46.3 & 61.11 & 82.91 & 21.80 \\
\bottomrule
 \end{tabularx}
\end{table}

\noindent\textbf{Modality Gap.}
Table~\ref{tab:overall_and_mc_combined} shows consistent drops on image-tagged items relative to text-only problems. For top-tier models, the gap from no-image to image conditions is typically 14--16 percentage points (e.g., 83.5\%\,$\to$\,68.2\% and 80.8\%\,$\to$\,66.9\%), and for mid-tier models it can approach ~20 points. This indicates that parsing and reasoning over structured visuals—graphs, grids, geometric diagrams—remain central bottlenecks, materially impacting overall accuracy.

\noindent\textbf{Standalone vs Among-Choices on MC.}
As discussed, we filter MC problems into choice-dependent and standalone subsets. Typically, standalone MC problems in competition settings are equipped with choices that appear correct at first sight to push students into making mistakes. Inspired by this, we propose an experimental setup to study this effect. For each model we compute: (i) \textbf{Standalone avg@8} = mean correctness over 8 samples; and (ii) \textbf{Among-Choices avg@8} = mean fraction of standalone problems whose final answer lies among the original options (not necessarily correct). This result can be seen in Table~\ref{tab:overall_and_mc_combined}. The large $\Delta$ values indicate that models consistently produce answers that coincide with some provided choice but not necessarily the correct one. This systematic gap between Among-Choices and Standalone accuracy reveals a susceptibility to these crafted competition traps. In other words, the trap choices often steers models toward distractor recognition rather than robust derivation, whereas the standalone format demands genuine solution construction. Moreover, the large $\Delta$ values provide strong support for adopting our evaluation suite as an RL environment, since models can potentially learn to avoid deliberately crafted distractors---an ability that is a prerequisite for performing well in competition-level reasoning.

\noindent\textbf{Topic-Level Performance.}
Per-topic accuracies highlight both broad strengths and persistent weaknesses. Top-tier models are strong in combinatorics, number reasoning, and invariants/monovariants, and they show competitive results in computational geometry (see Figure~\ref{tab:per_topic_accuracy} in Appendix~\ref{app:topic-scores}). In contrast, graph-theoretic subdomains (e.g., connectivity, matchings) and formal languages expose larger spreads across models, with mid-tier and lightweight/open-weight models struggling markedly. The dispersion suggests that discrete math reasoning is not uniformly mastered across mathematical domains.

\section{Conclusion}

Together, our findings indicate that CombiGraph-Vis yields strong separations across model families, exposes enduring multimodal reasoning deficits, and stresses the difference between distractor sensitive recognition and derivation-based solution. We leverage these observations in the discussion to analyze error modes and to outline methodological directions for building models that can
reliably solve complex, multimodal discrete mathematics problems.

\newpage
% \section{Conclusion} Together, our findings indicate that CombiGraph-Vis yields strong separations across model families, exposes enduring multimodal reasoning deficits, and stresses the difference between distractor-sensitive recognition and derivation-based solution. We leverage these observations in the Discussion to analyze error modes and to outline methodological directions for building models that can reliably solve complex, multimodal discrete mathematics problems.

% \section{LLM Usage Description}
% \label{app:llm-usage}
% We used LLMs such as gpt-5 and Gemini 2.5 Pro to polish writing, fix grammatical errors and latex alignment issues. 

\bibliographystyle{abbrvnat} % or plainnat/unsrtnat per NeurIPS style
\bibliography{AI4Math}       % <-- must match your uploaded .bib basename

\appendix

\section{Related Work}

\textbf{Mathematical Reasoning Benchmarks.}
GSM8K introduced 8,500 grade school math word problems with verification-based training, demonstrating that step-by-step solutions improve both accuracy and reliability\citep{cobbe2021training}. MATH scaled this approach to high school competition mathematics with 12,500 problems across algebra, geometry, number theory, and other domains\citep{hendrycks2021math}. Methodological advances complemented these datasets: chain-of-thought prompting enabled explicit reasoning steps\citep{wei2022chainofthought}, while self-consistency enhanced reliability through majority voting over multiple solution paths\citep{wang2023selfconsistency}. Competition-focused datasets followed with CHAMP providing 270 problems with rich concept-level annotations\citep{mao-etal-2024-champ} and OMNI-MATH aggregating 4,428 Olympiad-style problems from international competitions across over 33 mathematical sub-domains\citep{gao2024omnimath}.

\textbf{Visual Mathematical Reasoning.}
Visual mathematical reasoning benchmarks address problems where images contain essential information for solving mathematical questions. Domain-specific approaches include GeoQA with 5,010 geometric problems requiring diagram interpretation\citep{chen2021geoqa} and Conic10K with 10,861 conic section problems providing formal symbolic representations\citep{wu-etal-2023-conic10k}. Comprehensive collections followed: MathVista combines 6,141 visual math problems from 28 existing datasets spanning geometry, statistics, and algebraic reasoning\citep{lu2024mathvista}, MATH-V curates 3,040 competition problems requiring visual context understanding across 16 mathematical disciplines\citep{wang2024mathv}, and OlympiadBench extends beyond mathematics with 8,476 bilingual multimodal problems covering both mathematics and physics from international competitions\citep{he-etal-2024-olympiadbench}.
Compared to these, our benchmark centers discrete math style reasoning over graphs, grids, and combinatorial objects with short, checkable answers and technique labels.

\textbf{General Multimodal Reasoning.}
General multimodal reasoning benchmarks evaluate capabilities beyond mathematical domains. MMMU targets expert-level understanding with 11,500 college questions spanning art, business, science, health, humanities, and social science\citep{Yue_2024_CVPR}, while MMBench provides systematic evaluation across 20 ability dimensions with 3,000+ multiple-choice questions\citep{li2024mmbench}. Knowledge-intensive approaches include A-OKVQA with 25,000 questions requiring both visual understanding and world knowledge\citep{schwenk2022aokvqa} and CLEVR-Math with 10,000 synthetic questions testing systematic combination of arithmetic operations in visual contexts\citep{liu2022clevrmath}.

\textbf{Evaluation Methods and Robustness.}
Advanced evaluation methods examine solution quality and reasoning stability beyond final answer accuracy. We-Math introduces a diagnostic framework that decomposes 15,000 mathematical problems by knowledge concepts and evaluates models across four categories: insufficient knowledge, inadequate generalization, complete mastery, and rote memorization\citep{we-math-acl-2025}. DynaMath focuses on robustness evaluation by generating multiple variants of each seed problem, creating 501 base problems with over 5,000 variations to test consistency across input perturbations\citep{zou2025dynamath}, while MPBench provides a meta-evaluation framework for visual mathematical reasoning, testing models' abilities in step checking, solution aggregation, and guided step selection across 1,000 competition problems\citep{pan-etal-2025-mpbench}. Our evaluation complements these perspectives by quantifying modality gaps and distractor susceptibility (standalone vs. choice-dependent MC) in discrete, image-tagged settings.

\textbf{Solution Assessment.}
Evaluating open-ended mathematical solutions presents unique challenges requiring specialized assessment frameworks. HARP compiles 3,000 short-answer competition problems from prestigious contests, providing multiple human solution strategies and reference answers to enable comprehensive evaluation\citep{yue2024harp}, while U-MATH targets university-level mathematical reasoning with 1,100 problems spanning calculus, linear algebra, and advanced topics, introducing a meta-evaluation framework that assesses the quality of LLM-based grading systems\citep{chernyshev-etal-2025-u}. Tooling for automated answer verification supports reliable scoring of algebraic/numeric responses (e.g., Math-Verify)\citep{kydlicek2024mathverify}. Further, reducing proof-based tasks to final-answer grading can misalign with intended assessment goals\citep{mahdavi2025brains}. CombiGraph-Vis focuses on short, checkable formats paired with verified solutions and reports results by format and modality to align evaluation with task intent.

\section{Task Formats and Verification Protocol}
We evaluate models by generating eight solutions per problem using a chain-of-thought prompt that instructs models to produce step-by-step reasoning and wrap the final answer in \texttt{\textbackslash boxed\{\}} format (Appendix~\ref{app:solution_prompt}). For choice-dependent multiple-choice problems, we include the answer choices in the prompt to ensure the model selects from the provided options. To parse the the final answer from the model's output, we use a simple regex pattern that matches the \texttt{\textbackslash boxed\{\}} format. If all of the choices for that specific problem were numerical/algebraic expressions, we used the Math-Verify\cite{kydlicek2024mathverify} library to check if the extracted answer is equivalent to the final answer. In case the generated solution didn't follow the instruction and didn't wrap the final answer in \texttt{\textbackslash boxed\{\}}, or the choices were not numerical/algebraic expressions, we offloaded the task to an LLM (Gemini 2.5 Flash) to extract the final answer. In the prompt, we asked the model to extract the final answer's raw value, and the matching choice (if any) and the standardized form of the final answer (in case the choices were not numerical/algebraic expressions and the final answer matched one of the choices). We then checked if the extracted answer is equal to the final answer or the extracted choice is equal to the correct option.

\section{Technique Labels and Taxonomy}
To enable fine-grained analysis of mathematical reasoning capabilities, we applied technique labeling based on the official Iranian Informatics Olympiad curriculum. Each problem receives hierarchical labels following a three-level taxonomy: Topic $\rightarrow$ Sub-topic $\rightarrow$ Sub-sub-topic (e.g., Combinatorics $\rightarrow$ Counting Foundations $\rightarrow$ Stars \& bars). We use a single prompt that assigns labels based on techniques that explicitly appear in solution steps. The taxonomy covers 13 major topics spanning discrete mathematics with 89 distinct sub-sub-topic labels that capture precise mathematical approaches used in solutions. This fine-grained labeling enables researchers to analyze model performance across specific techniques, identify capability gaps, and design targeted evaluation protocols. The complete hierarchical taxonomy and labeling prompt are provided in Appendix~\ref{app:hier-taxonomy}.

\section{Topic Level Performance}
\label{app:topic-scores}
\begin{figure}[htbp]
    \centering
    \includegraphics[width=0.98\linewidth]{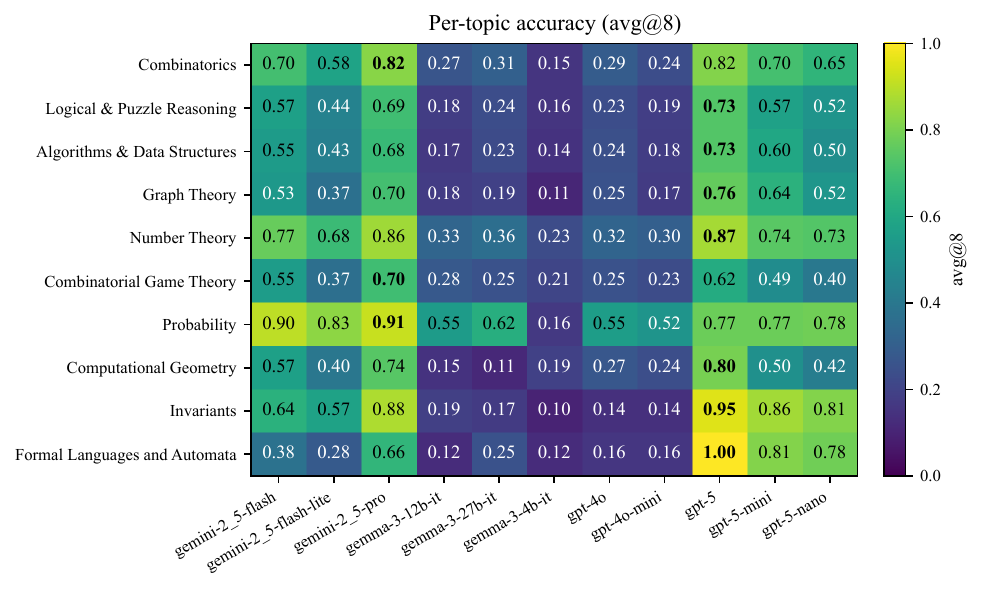}
    \caption{Per-model accuracy by topic (\%). Best score per topic is highlighted in bold within each cell.}
    \label{tab:per_topic_accuracy}
\end{figure}

\section{Full Examples for Problem Categories}
\label{app:examples}

\begin{tcolorbox}[
    enhanced,breakable,
    colback=lightcream,
    colframe=mediumblue,
    colbacktitle=darkercream,
    coltitle=black,
    title={\textbf{Choice-Dependent Problem (dataset exemplar)}}
]
A calculating machine has an internal memory called $M$. This machine can calculate an expression by performing the following instructions:
\begin{itemize}
    \item \texttt{Add X}: Adds the value of $X$ to the value of $M$ and stores the result in $M$.
    \item \texttt{Mul X}: Multiplies the value of $X$ by the value of $M$ and stores the result in $M$.
\end{itemize}
In the above instructions, $X$ can be an integer or a variable. Assume the initial value of $M$ is zero. For example, the following instructions, from left to right, calculate the expression $ax+5$: \texttt{Add a}, \texttt{Mul x}, \texttt{Add 5}. \textbf{Which of the following expressions cannot be calculated by this machine?}

\vspace{0.2cm}
\begin{enumerate}
    \item $ax^2+bx+c$
    \item $(a+b)xy+ya$
    \item $(ax+by)(a+b)$
    \item $3x^5 + 1$
    \item All these expressions can be calculated
\end{enumerate}
\end{tcolorbox}

\vspace{0.3cm}

\begin{tcolorbox}[
    enhanced,breakable,
    colback=warmbeige,
    colframe=mediumblue,
    colbacktitle=darkerbeige,
    coltitle=black,
    title={\textbf{Standalone Problem (dataset exemplar)}}
]
We have written numbers $1$ to $78$ clockwise on a circle. We select the number $1$ as the current number and repeat the following operations until only one number remains on the circle:
\begin{itemize}
    \item If the current number is $x$, remove it from the circle, add one unit to the $x$ next numbers clockwise on the circle, and select the number after that (two places clockwise from the removed number) as the current number.
\end{itemize}

Note that if the number of remaining numbers on the circle is less than $3$, one or more numbers might have more than one unit added to them. 

\textbf{What is the remainder when the number that finally remains on the circle is divided by $5$?}

\vspace{0.2cm}
\begin{enumerate}
    \item $0$
    \item $1$
    \item $2$
    \item $3$
    \item $4$
\end{enumerate}
\end{tcolorbox}

\begin{tcolorbox}[
    enhanced,breakable,
    colback=lightblue,
    colframe=mediumblue,
    colbacktitle=softblue,
    coltitle=black,
    title={\textbf{Context-Dependent Problem (dataset exemplar)}}
]
\textbf{Context:} Consider the following definition for the next three questions: An $m \times n$ table where each cell contains an integer is called a 'counting table' if the absolute difference of the numbers written in any two adjacent (row-wise or column-wise) cells is exactly one. As an example, the table below is a $2 \times 3$ counting table.

\vspace{0.2cm}
\begin{center}
\IfFileExists{figures/img-3_svg-tex.pdf}{%
\includesvg[width=0.4\textwidth]{figures/img-3.svg}
}{}
\end{center}

% Fallback table representation
\begin{center}
\begin{tabular}{|c|c|c|}
\hline
2 & 3 & 2 \\
\hline
3 & 2 & 1 \\
\hline
\end{tabular}
\end{center}
\vspace{0.2cm}

\textbf{Question:} A counting $m \times n$ table, with all its cells filled, is given. We want to reveal the numbers in a minimum number of its cells (their numbers become known to us) so that we can deduce the numbers in the remaining cells. In what range does this minimum lie?

\vspace{0.2cm}
\begin{enumerate}
    \item 1 or 2
    \item $[3, m+n-1]$
    \item $[\frac{mn}{2}, m+n]$
    \item $[\frac{mn}{2}, mn-1]$
    \item Exactly $mn$
\end{enumerate}
\end{tcolorbox}

\section{Dataset Statistics}
\begin{table}[H]
\begin{center}
\begin{tabular}{lrrr}
\bf Category & \bf Count & \bf \% of Total & \bf With Images \\
\hline
\textbf{All Problems} & \textbf{1135} & \textbf{100.0} & \textbf{406 (35.8\%)} \\
\hline
Short-answer & 884 & 77.9 & 321 (36.3\%) \\
Multiple-choice & 157 & 13.8 & 49 (31.2\%) \\
Yes/No & 94 & 8.3 & 36 (38.3\%) \\
\end{tabular}
\end{center}
\caption{CombiGraph-Vis dataset statistics.}
\label{tab:dataset_overview}
\end{table}

\section{Deferred Definitions and Explanations}
\label{app:severity_definitions}
\noindent\textbf{Overall Error Severity.}
\label{app:severity_definitions_taxonomy}
\begin{enumerate}
  \item \textbf{No issues.} Clear, correct; e.g., punctuation/spacing only.
  \item \textbf{Minor issues.} Small typos/notation/wording that do not change interpretation; e.g., $\binom{n}{k}$ written as $C(n,k)$.
  \item \textbf{Moderate issues.} Multiple minor issues or one ambiguity; intended reading still recoverable; e.g., missing variable domain but inferable.
  \item \textbf{Major issues.} Contradiction, missing crucial data, or a flawed step that invalidates the solution path; e.g., incorrect identity used.
  \item \textbf{Critical failure.} Pervasive/fatal problems (nonsense/corrupted content or irreconcilable mismatch); e.g., unreadable required figure or answer contradicts solution.
\end{enumerate}

\noindent\textbf{Error Category Definitions.}
\begin{enumerate}
  \item \textbf{Conversion errors.} Parser/OCR defects (notation, formatting, encoding); e.g., $\binom{n}{k}\!\to\!\tfrac{n}{k}$, dropped subscripts.
  \item \textbf{Translator/annotator errors.} Translation or metadata mistakes (typos, choice permutation, mislabel); e.g., swapped options B/C.
  \item \textbf{Original source errors.} Issues in archived materials/errata; e.g., mis-scanned digit or incorrect constant in the source PDF.
\end{enumerate}

\noindent\textbf{Critic Definitions.}
\label{app:critics_definitions}
\begin{enumerate}
  \item \textbf{Typo/Clarity Critic.} Flags typos, translation slips, and clarity/formatting issues by comparing English with source text; e.g., inconsistent notation or mistranslated term.
  \item \textbf{Logical Soundness Critic.} Checks step-by-step reasoning and computations; e.g., unjustified inequality step or omitted counting case.
  \item \textbf{Final Answer Match.} Makes sure that the final answer stated in the solution matches the final answer stored in the database.
\end{enumerate}

\section{Implementation Details}
\label{app:implementation}

\begin{algorithm}[htbp]
\caption{Problem Validation Workflow (First Phase)}
\label{alg:problem_validation}
\begin{algorithmic}[1]
\Require Problem datum $d=$ (problem, choices, english\_solution, context, correct\_option, answer\_value, crawled\_persian\_markdown, svg\_sources)
\Ensure problem\_validation\_data

\State reports $\gets$ [ ]
\For{$i \gets 1$ to $3$}
  \State typo\_report $\gets$ TypoClarityCritic($d$)
  \State logic\_report $\gets$ LogicalSoundnessCritic($d$)
  \State answer\_report $\gets$ AnswerVerificationCritic($d$)
  \State combined\_report $\gets$ ReportCollector(typo\_report, logic\_report, answer\_report)
  \State Append(reports, combined\_report)
\EndFor

\State joined\_reports $\gets$ JoinReportChunks(reports)
\State validation\_result $\gets$ FinalAggregator(joined\_reports)

\State \Return validation\_result
\end{algorithmic}
\end{algorithm}

\begin{algorithm}[htbp]
\caption{Error Detection and Classification}
\label{alg:error_detection}
\begin{algorithmic}[1]
\Require Problem datum $d=$ (problem, choices, english\_solution, context, correct\_option, answer\_value, crawled\_persian\_markdown, svg\_sources)
\Ensure Classification result $agg$ with fix requirements

\State findings\_md $\gets$ BuildFindingsText(LoadValidationData($d$.id))

\State reports $\gets$ [ ]
\For{$i \gets 1$ to $3$}
  \State $r \gets$ IssueDetector($d$, findings\_md)
  \State Append(reports, $r$)
\EndFor
\State reports\_md $\gets$ JoinIssueReportChunks(reports)

\State agg $\gets$ IssueAggregator(reports\_md, $d$)

\If{agg.is\_original\_source\_error}
  \State engagement\_md $\gets$ SolutionEngager($d$, agg.aggregated\_report\_md)
  \State src\_cls $\gets$ IssueDetectorWithEngagement($d$, engagement\_md)
  \State src\_cls\_md $\gets$ FormatToMarkdown(src\_cls)
  \State agg $\gets$ EngagementReportSynthesizer(agg.aggregated\_report\_md, engagement\_md, src\_cls\_md)
  \If{agg.requires\_human\_intervention}
    \State \Return ComposeHumanInterventionReport(agg)
  \EndIf
\ElsIf{agg.is\_image\_understanding\_issue}
  \State \Return ComposeHumanInterventionReport(agg)
\EndIf

\State \Return agg \Comment{Classification result for automated fixing}
\end{algorithmic}
\end{algorithm}

\begin{algorithm}[htbp]
\caption{Automated Error Resolution and Fixing}
\label{alg:automated_fixing}
\begin{algorithmic}[1]
\Require Problem datum $d$, classification result $agg$ from Algorithm~\ref{alg:error_detection}
\Ensure Fixed problem data or human intervention report

\State fix\_plan\_md $\gets$ FixPlanner(agg.aggregated\_report\_md, $d$)
\State fixed $\gets$ Fixer(fix\_plan\_md, $d$)
\State ctx $\gets$ UpdateContextWithFixes(fixed)
\State fixed\_md $\gets$ FormatFixedData(ctx.fixed\_problem\_data)

\State successes $\gets 0$
\For{$t \gets 1$ to $20$}
  \State result $\gets$ Validator(agg.aggregated\_report\_md, fix\_plan\_md, $d$, fixed\_md)
  \If{result.is\_fixed}
    \State successes $\gets$ successes $+ 1$
    \If{successes $\ge 5$}
      \State \textbf{break}
    \EndIf
  \Else
    \State successes $\gets 0$
    \State fix\_plan\_md $\gets$ RePlanner(agg.aggregated\_report\_md, result.reasoning, fix\_plan\_md, $d$)
    \State fixed $\gets$ Fixer(fix\_plan\_md, $d$)
    \State ctx $\gets$ UpdateContextWithFixes(fixed)
    \State fixed\_md $\gets$ FormatFixedData(ctx.fixed\_problem\_data)
  \EndIf
\EndFor

\State \Return ComposeAutoFixOutput($d$, agg, fix\_plan\_md, fixed\_md)
\end{algorithmic}
\end{algorithm}
\FloatBarrier
\section{Prompt Specifications}

\subsection{Problem Validation Prompts}

\subsubsection{TypoClarityCritic}
\ShowMD{TypoClarityCritic Prompt}{Prompts/problem_validation/typo_clarity_critic_prompt.md}

\subsubsection{LogicalSoundnessCritic}
\ShowMD{LogicalSoundnessCritic Prompt}{Prompts/problem_validation/logical_soundness_critic_prompt.md}

\subsubsection{AnswerVerificationCritic}
\ShowMD{AnswerVerificationCritic Prompt}{Prompts/problem_validation/answer_verification_critic_prompt.md}

\subsubsection{FinalAggregator}
\ShowMD{FinalAggregator Prompt}{Prompts/problem_validation/final_aggregator_prompt.md}

\subsection{Error Resolution Prompts}

\subsubsection{IssueDetector}
\ShowMD{IssueDetector Prompt}{Prompts/problem_issue_resolver/issue_detector_prompt.md}

\subsubsection{IssueAggregator}
\ShowMD{IssueAggregator Prompt}{Prompts/problem_issue_resolver/issue_aggregator_prompt.md}

\subsubsection{SolutionEngager}
\ShowMD{SolutionEngager Prompt}{Prompts/problem_issue_resolver/solution_engager_prompt.md}

\subsubsection{IssueDetectorWithEngagement}
\ShowMD{IssueDetectorWithEngagement Prompt}{Prompts/problem_issue_resolver/issue_detector_with_engagement_prompt.md}

\subsubsection{EngagementReportSynthesizer}
\ShowMD{EngagementReportSynthesizer Prompt}{Prompts/problem_issue_resolver/engagement_report_synthesizer_prompt.md}

\subsubsection{FixPlanner}
\ShowMD{FixPlanner Prompt}{Prompts/problem_issue_resolver/fix_planner_prompt.md}

\subsubsection{Fixer}
\ShowMD{Fixer Prompt}{Prompts/problem_issue_resolver/fixer_prompt.md}

\subsubsection{Validator}
\ShowMD{Validator Prompt}{Prompts/problem_issue_resolver/validator_prompt.md}

\subsubsection{RePlanner}
\ShowMD{RePlanner Prompt}{Prompts/problem_issue_resolver/re_planner_prompt.md}

\section{Complete Technique Taxonomy}
\label{app:taxonomy}

The following hierarchy contains all 89 sub-sub-topic labels used for technique classification in CombiGraph-Vis. Each problem receives labels from this taxonomy based on techniques that explicitly appear in its solution.

\subsection{Technique Labeling Prompt}

\ShowMD{Technique Labeler Prompt}{Prompts/technique_labeler/technique_labeler.md}

\subsection{Solution Generation Prompt}
\label{app:solution_prompt}

\ShowMD{Solution Generation Prompt}{Prompts/solver/solution_generation.md}

\subsection{Hierarchical Taxonomy of Topics in CombiGraph-Vis}
\label{app:hier-taxonomy}

\begingroup\small
\input{taxonomy_trees_fragment.tex}
\endgroup

\end{document}

%% file: math_commands.tex
\usepackage{amsmath,amsfonts,bm}

\def\eqref#1{equation~\ref{#1}}
\def\1{\bm{1}}

\DeclareMathAlphabet{\mathsfit}{\encodingdefault}{\sfdefault}{m}{sl}
\SetMathAlphabet{\mathsfit}{bold}{\encodingdefault}{\sfdefault}{bx}{n}

%% file: taxonomy_trees_fragment.tex
% --------- BEGIN: Taxonomy Trees Fragment (no preamble) ---------
% Assumes neurips_2024 style already loaded hyperref; we don't reload it.
% Required packages in your preamble:
% \usepackage{xcolor}
% \usepackage[edges]{forest}
% \usepackage{tikz}
% \usetikzlibrary{arrows.meta,positioning}

% ---- Colors (domains) ----
\providecolor{domA}{HTML}{1E88E5} % Combinatorics
\providecolor{domB}{HTML}{8E24AA} % Graph Theory
\providecolor{domC}{HTML}{F4511E} % Game Theory
\providecolor{domD}{HTML}{43A047} % Probability
\providecolor{domE}{HTML}{FB8C00} % Number Theory
\providecolor{domF}{HTML}{3949AB} % FL&A
\providecolor{domG}{HTML}{0097A7} % Algorithms
\providecolor{domH}{HTML}{6D4C41} % Data Structures
\providecolor{domI}{HTML}{7CB342} % Strings
\providecolor{domJ}{HTML}{5E35B1} % Discrete Geometry
\providecolor{domK}{HTML}{00897B} % Logic & Puzzles
\providecolor{domL}{HTML}{C2185B} % Inequalities
\providecolor{domM}{HTML}{546E7A} % Proof Strategies

% ---- Forest styles ----
\forestset{
  taxonomy/.style={
    for tree={
      grow'=0,
      parent anchor=east,
      child anchor=west,
      align=left,
      s sep=5pt,
      l sep=10pt,
      inner sep=2pt,
      edge path'/.style={draw, ->, >={Stealth}},
      if level=0{font=\large\bfseries}{}
    }
  },
  lvl1/.style={font=\bfseries},
  leaf/.style={},
}
\newcommand{\domainlabel}[2]{\noindent{\large\bfseries\textcolor{#1}{#2}}\par\vspace{0.25em}}

% ---- Trees ----
{\small

% ---------------- Combinatorics ----------------
\domainlabel{domA}{Combinatorics}
\begin{forest} taxonomy
[Combinatorics, draw, rounded corners, fill=domA!6
  [Counting Foundations, lvl1
    [Sum/Product/Complement, leaf]
    [Bijections, leaf]
    [Permutations, leaf]
    [Combinations, leaf]
    [Stars \& Bars, leaf]
    [Binomial thm.; lattice paths; identities, leaf]
  ]
  [Advanced Counting, lvl1
    [Inclusion--Exclusion, leaf]
    [Double counting, leaf]
  ]
  [Recurrences \& GFs, lvl1
    [Linear recurrences, leaf]
    [{Classic sequences (Fib., Catalan)}, leaf]
    [Light OGFs/EGFs, leaf]
  ]
  [Symmetry Counting, lvl1
    [Burnside's lemma, leaf]
    [Pólya enumeration (intro), leaf]
  ]
  [Invariants, lvl1
    [Parity/modular invariants, leaf]
    [Coloring/weighting, leaf]
    [Termination via monovariants, leaf]
  ]
  [Probabilistic Method, lvl1
    [Linearity-of-expectation, leaf]
    [Existence via expectation, leaf]
  ]
]
\end{forest}

% ---------------- Graph Theory ----------------
\domainlabel{domB}{Graph Theory}
\begin{forest} taxonomy
[Graph Theory, draw, rounded corners, fill=domB!6
  [Basics, lvl1
    [Defs; adjacency list/matrix, leaf]
    [Degree/handshaking; graphic seq., leaf]
    [Isomorphism; BFS/DFS; paths/cycles/dist., leaf]
  ]
  [Trees, lvl1
    [Props; rooted/binary trees, leaf]
    [DFS/BFS trees, leaf]
    [Spanning trees \& counting, leaf]
  ]
  [Connectivity, lvl1
    [Connectedness; cut vertices/bridges, leaf]
    [k-connectivity; blocks (biconnected), leaf]
  ]
  [Directed Graphs, lvl1
    [Strongly connected comps., leaf]
    [Tournaments, leaf]
  ]
  [Cycles \& Trails, lvl1
    [Eulerian trails/tours, leaf]
    [Hamiltonian paths/cycles, leaf]
  ]
  [Matchings \& Covers, lvl1
    [Bipartite matchings; Hall, leaf]
    [General matchings; independence no., leaf]
    [Vertex/edge covers; bipartite relations, leaf]
  ]
  [Planarity \& Coloring, lvl1
    [Planar; Euler's formula (apps), leaf]
    [Vertex/edge coloring; counting colorings, leaf]
  ]
]
\end{forest}

% ---------------- Combinatorial Game Theory ----------------
\domainlabel{domC}{Combinatorial Game Theory}
\begin{forest} taxonomy
[Comb. Game Theory, draw, rounded corners, fill=domC!6
  [Modeling \& State Analysis, lvl1
    [Game graphs; W/L/D states, leaf]
    [DP for evaluation; kernels; strategy existence, leaf]
  ]
  [Canonical Examples, lvl1
    [Nim; partisan games; Hex; Shannon switching, leaf]
  ]
]
\end{forest}

% ---------------- Probability (Elementary) ----------------
\domainlabel{domD}{Probability}
\begin{forest} taxonomy
[Probability, draw, rounded corners, fill=domD!6
  [Core Concepts, lvl1
    [Sample spaces \& events; basic prob., leaf]
    [Conditional prob.; independence; Bernoulli, leaf]
  ]
  [Expectation, lvl1
    [RV; linearity of expectation, leaf]
    [Indicator variables, leaf]
  ]
]
\end{forest}

% ---------------- Number Theory ----------------
\domainlabel{domE}{Number Theory (Contest Essentials)}
\begin{forest} taxonomy
[Number Theory, draw, rounded corners, fill=domE!6
  [Divisibility \& GCD/LCM, lvl1
    [Euclidean alg.; B\'ezout, leaf]
  ]
  [Primes \& Congruences, lvl1
    [Modular arithmetic; FLT; CRT, leaf]
  ]
  [Counting Toolbox, lvl1
	[{$\tau,\ \sigma,\ \varphi$; multiplicativity}, leaf]
    [Fast exp.; modular inverses, leaf]
    [Counts via gcd/lcm; CRT-based counts, leaf]
  ]
]
\end{forest}

% ---------------- Formal Languages & Automata ----------------
\domainlabel{domF}{Formal Languages \& Automata}
\begin{forest} taxonomy
[FL \& Automata, draw, rounded corners, fill=domF!6
  [Languages, lvl1
    [{Alphabets, strings, languages}, leaf]
  ]
  [Machines, lvl1
    [DFA \& NFA; pushdown automata; Turing machines, leaf]
  ]
]
\end{forest}

% ---------------- Algorithmic Techniques ----------------
\domainlabel{domG}{Algorithmic Techniques}
\begin{forest} taxonomy
[Algorithmic Techniques, draw, rounded corners, fill=domG!6
  [Greedy, lvl1
    [Exchange arguments; counterexample design, leaf]
  ]
  [DP, lvl1
    [State modeling, leaf]
  ]
  [Recursion, lvl1
    [Recurrences; correctness ideas, leaf]
  ]
  [Search, lvl1
    [Backtracking \& pruning; BFS/DFS patterns, leaf]
  ]
  [Classic Tricks, lvl1
    [Binary search on answer; two-pointers/sliding window, leaf]
  ]
  [Proof of Correctness, lvl1
    [Invariants; loop/phase arguments, leaf]
  ]
]
\end{forest}

% ---------------- Conceptual Data Structures ----------------
\domainlabel{domH}{Conceptual Data Structures}
\begin{forest} taxonomy
[Data Structures, draw, rounded corners, fill=domH!6
  [Linear Containers, lvl1
    [{Stack, queue, deque}, leaf]
  ]
  [Priority \& Set Structures, lvl1
    [Heaps; sets/maps; hashing ideas, leaf]
  ]
  [Disjoint Set Union, lvl1
    [Connectivity; cycle detection, leaf]
  ]
  [Graph Representations, lvl1
    [Adjacency list vs matrix; trade-offs, leaf]
  ]
]
\end{forest}

% ---------------- Strings & Combinatorics on Words ----------------
\domainlabel{domI}{Strings \& Combinatorics on Words}
\begin{forest} taxonomy
[Strings \& Words, draw, rounded corners, fill=domI!6
  [Structural Properties, lvl1
    [Prefix/suffix/border; periodicity, leaf]
    [Palindromes, leaf]
  ]
  [Counting \& Constraints, lvl1
    [Counting constrained strings, leaf]
    [Links to automata (acceptance as constraints), leaf]
  ]
]
\end{forest}

% ---------------- Discrete & Computational Geometry ----------------
\domainlabel{domJ}{Computational Geometry}
\begin{forest} taxonomy
[Computational Geometry, draw, rounded corners, fill=domJ!6
  [Primitives, lvl1
    [Orientation test, leaf]
    [Line/segment intersection, leaf]
  ]
  [Polygons \& Lattice, lvl1
    [Polygon area, leaf]
    [Lattice points; Pick's theorem, leaf]
  ]
  [Convexity, lvl1
    [Convex hull intuition \& uses, leaf]
  ]
]
\end{forest}

% ---------------- Logical & Puzzle Reasoning ----------------
\domainlabel{domK}{Logical \& Puzzle Reasoning}
\begin{forest} taxonomy
[Logic \& Puzzle Reasoning, draw, rounded corners, fill=domK!6
  [Logic \& Proof Moves, lvl1
    [Propositional logic; contradiction/contrapositive, leaf]
  ]
  [Puzzle Tactics, lvl1
    [Invariants for grids/tilings; parity tricks, leaf]
    [Constructive examples \& counterexamples, leaf]
  ]
]
\end{forest}

% ---------------- Inequalities & Algebraic Tools ----------------
\domainlabel{domL}{Inequalities \& Algebraic Tools}
\begin{forest} taxonomy
[Inequalities \& Algebraic Tools, draw, rounded corners, fill=domL!6
  [Core Inequalities, lvl1
    [AM--GM; Cauchy--Schwarz (incl. Titu), leaf]
    [Rearrangement inequality, leaf]
  ]
  [Summation Tricks, lvl1
    [Telescoping; bounding techniques, leaf]
  ]
]
\end{forest}

% ---------------- General Proof Strategies ----------------
\domainlabel{domM}{Proof Strategies}
\begin{forest} taxonomy
[Proof Strategies, draw, rounded corners, fill=domM!6
  [Induction, lvl1
    [Weak/Strong; structural, leaf]
    [Strengthening hypotheses; infinite descent/minimal counterexample, leaf]
  ]
  [Pigeonhole Principle, lvl1
    [Simple; generalized/strong , leaf]
    [{Apps: geometry, number theory, graphs}, leaf]
  ]
  [Extremal Principle, lvl1
    [Max/Min arguments; extremal objects, leaf]
  ]
  [Coloring \& Invariants, lvl1
    [Checkerboard/parity coloring, leaf]
    [Invariants \& monovariants, leaf]
  ]
]
\end{forest}

} % end \small

% --------- END: Taxonomy Trees Fragment ---------